\documentclass[runningheads]{llncs}
\usepackage[T1]{fontenc}
\usepackage{graphicx,verbatim}
\usepackage{xcolor}
\usepackage{xspace}
\usepackage{amsmath,amsfonts,amssymb}
\usepackage{booktabs}
\usepackage{multirow}
\usepackage{hyperref}

\makeatletter
\DeclareRobustCommand\onedot{\futurelet\@let@token\@onedot}
\def\@onedot{\ifx\@let@token.\else.\null\fi\xspace}
\def\eg{\emph{e.g}\onedot}

\begin{document}
\title{OSCAR: Occupancy-based Shape Completion via Acoustic Neural Implicit Representations}
\begin{comment}  %% Removed for anonymized MICCAI submission
\author{First Author\inst{1}\orcidID{0000-1111-2222-3333} \and
Second Author\inst{2,3}\orcidID{1111-2222-3333-4444} \and
Third Author\inst{3}\orcidID{2222--3333-4444-5555}}
%
\authorrunning{F. Author et al.}
\institute{Princeton University, Princeton NJ 08544, USA \and
Springer Heidelberg, Tiergartenstr. 17, 69121 Heidelberg, Germany
\email{lncs@springer.com}\\
\url{http://www.springer.com/gp/computer-science/lncs} \and
ABC Institute, Rupert-Karls-University Heidelberg, Heidelberg, Germany\\
\email{\{abc,lncs\}@uni-heidelberg.de}}

\end{comment}

\author{Magdalena Wysocki\inst{1, 2}\thanks{These authors contributed equally to this work.}\and
Kadir Burak Buldu\inst{1}\textsuperscript{*}
 \and
 Miruna-Alexandra Gafencu\inst{1, 2, 3}
 \and
 Benjamin D. Killeen \inst{1, 2}
 \and
 Mohammad F. Azampour\inst{1, 2}\and
Nassir Navab\inst{1, 2} }
%index{Wysocki, Magdalena}
%index{Buldu, Kadir Burak}
%index{Gafencu, Miruna-Alexandra}
%index{Killeen, Benjamin D.}
%index{Azampour, Mohammad F.}
%index{Navab, Nassir}
%
\authorrunning{M. Wysocki et al.}
% Corresponding author: Magdalena Wysocki (magdalena.wysocki@tum.de)
\institute{Chair for Computer Aided Medical Procedures (CAMP) \\ Technical University of Munich, Boltzmannstr. 3, 85748 Garching, Germany 
\email{\{magdalena.wysocki, burak.buldu, miruna.gafencu, bd.killeen, mf.azampour, nassir.navab\}@tum.de}
 \and
Munich Center for Machine Learning (MCML), Munich, Germany
\and Konrad Zuse School of Excellence in Reliable AI (relAI), Germany}
% \author{Anonymized Authors}  %% Added for anonymized MICCAI submission
% \authorrunning{Anonymized Author et al.}
% \institute{Anonymized Affiliations \\
%     \email{email@anonymized.com}}
  
\maketitle              % typeset the header of the contribution
\begin{abstract}
Accurate 3D reconstruction of vertebral anatomy from ultrasound is important for guiding minimally invasive spine interventions, but it remains challenging due to acoustic shadowing and view-dependent signal variations. We propose an occupancy-based shape completion method that reconstructs complete 3D anatomical geometry from partial ultrasound observations.
Crucially for intra-operative applications, our approach extracts the anatomical surface directly from the image, avoiding the need for anatomical labels during inference.
This label-free completion relies on a coupled latent space representing both the image appearance and the underlying anatomical shape. 
By leveraging a Neural Implicit Representation (NIR) that jointly models both spatial occupancy and acoustic interactions, the method uses acoustic parameters to become implicitly aware of the unseen regions without explicit shadowing labels through tracking acoustic signal transmission.
We show that this method outperforms state-of-the-art shape completion for B-mode ultrasound by 80\% in HD95 score.
We validate our approach both \emph{in silico} and on phantom US images with registered mesh models from CT labels, demonstrating accurate reconstruction of occluded anatomy and robust generalization across diverse imaging conditions.
Code and data will be released on publication.

% Authors must provide keywords and are not allowed to remove this Keyword section.
\end{abstract}
%
%
% yielding only fragmented views of the bony anatomy.  Relying solely on this visible information restricts visualization to the bone surface. 
\section{Introduction}
Ultrasound (US) image-guided spinal surgery offers significant benefits for both patients and clinicians, avoiding excessive ionizing radiation while retaining the benefits of minimally invasive approaches. Unfortunately, severe acoustic shadowing and reverberation significantly degrades image quality, so that only the near surface of hard tissue can be directly observed. To provide intra-operative guidance, then, expert sonographers rely on strong prior knowledge of spine anatomy to acquire multiple viewpoints and mentally infer the complete 3D geometry. Anatomical shape completion mimics this cognitive process, moving beyond visible surface reconstruction~\cite{chen2024rocosdf,wysocki2025ultron} to create the anatomical digital twin, \eg, for robotic- and computer-assisted approaches transitioning from distinct (visible) to completed (occluded)~\cite{duque2024ultrasound} anatomical reconstruction.
In the past, shape completion methods have used statistical shape models to approximate this expert capability~\cite{gafencu2024shape,gafencu2025us,massalimova2025surgpointtransformer}, but the reliance on explicitly annotated, discrete point clouds inherently ignores the acoustic image formation process, treating shadows as missing data and ignoring their view-dependent properties.
% Here, we propose a physics-informed shape completion framework that leverages both anatomical and acoustic priors to reconstruct the complete vertebral geometry directly from fragmented B-mode observations, thereby improving 3D spatial guidance for minimally invasive spine interventions and robotic surgery.
Our goal is to provide robust, physics-informed shape completion by leveraging both anatomical and acoustic priors to reconstruct the complete vertebral geometry from fragmented B-mode images, thereby improving 3D spatial guidance for minimally invasive spine interventions and robotic surgery.
%  Statistical Shape Models (SSMs)US4, 5 or by utilizing voxel-based 3D deep learning arrays [cite]SSMs are geometrically rigid and strictly constrained by their training distributions, while discrete voxel grids suffer from severe resolution constraints and high memory demands. both

% - ultrasound data are generally hard to obtain, especially matching pairs with completed shapes\
% - one can use simulations to get population data for pretraining

Recently, Neural Implicit Representations (NIRs) have emerged as a powerful alternative to discrete surfaces for complex shape modeling.
Foundational works in computer vision, such as DeepSDF~\cite{park2019deepsdf} and Occupancy Networks~\cite{mescheder2019occupancy}, demonstrated that NIRs can effectively encode complex 3D shapes. 
Within medical imaging, approaches like ImplicitAtlas~\cite{yang2022implicitatlas}, alongside other recent implicit shape models~\cite{de2025steerable,yang2024generating} have further shown that these networks can learn rich, continuous shape priors, while shared latent spaces encode both anatomical shape and image appearance~\cite{stolt2023nisf,vyas2025fit}.
% Building upon these geometric priors, medical imaging NIR-based segmentation frameworks~\cite{stolt2023nisf, vyas2025fit} have shown that NIRs can learn a rich, shared latent space coupling anatomical shape and image appearance. 
By training on both images and anatomical segmentations, these methods enable self-supervised test-time optimization (TTO), where the shared latent space is optimized using only a medical image to accurately infer the segmentation. Thus far, however, NIRs with coupled latent spaces have been ill-suited to the task of shape completion in ultrasound due to the severe occlusions of acoustic shadowing and view-dependent characteristics of B-mode images. Without a modality-specific model, incomplete and contradicting information from multiple views corrupts the reconstruction process.

% Moving beyond standard segmentation tasks, we are the first to demonstrate that NIRs with the coupled latent space are highly suitable for complex shape completion in ultrasound.
% However, applying these frameworks to ultrasound is an inherently ill-defined problem as severe acoustic shadowing leaves large portions of the vertebral anatomy invisible. Furthermore, in multi-view ultrasound, view-dependent shadowing naturally results in contradicting B-mode intensities for the exact same physical location. Consequently, matching the underlying geometry with raw pixel intensities becomes highly ambiguous; without physical context, forcing a network to reconcile these contradictions severely degrades the reconstructed geometry.

Here, we present Occupancy-based Shape Completion via Acoustic NIRs (OSCAR), a physics-aware framework for US-based spinal shape completion that integrates view-dependent acoustic properties of tissue into a coupled-latent NIR.
By modeling acoustic image formation and signal propagation as a ray-tracing process in this NIR~\cite{burger2012real,wysocki2024ultra,salehi2015patient}, we enable the latent space to resolve multi-view pixel contradictions and reconstruct unobserved regions with information from disparate viewpoints.
Further, this coupling facilitates bidirectional optimization. Given an inferred anatomical shape, our model can reconstruct a plausible acoustic space, allowing for novel view synthesis of B-mode images with known shape.
We evaluate our approach both \emph{in silico} and on real images of an anthropomorphic phantom, with ground truth from registered computed tomography (CT) imaging. 
Compared to the state-of-the-art in ultrasound shape completion SITD~\cite{gafencu2024shape}, OSCAR improves the HD95 score of reconstructed shapes by 80\% while maintaining anatomically plausible shapes across its latent space, in addition to enabling realistic image synthesis from known shape.

\section{Method}
\begin{figure}[htbp] % h=here, t=top, b=bottom, p=page
    \centering
    \includegraphics[width=\textwidth]{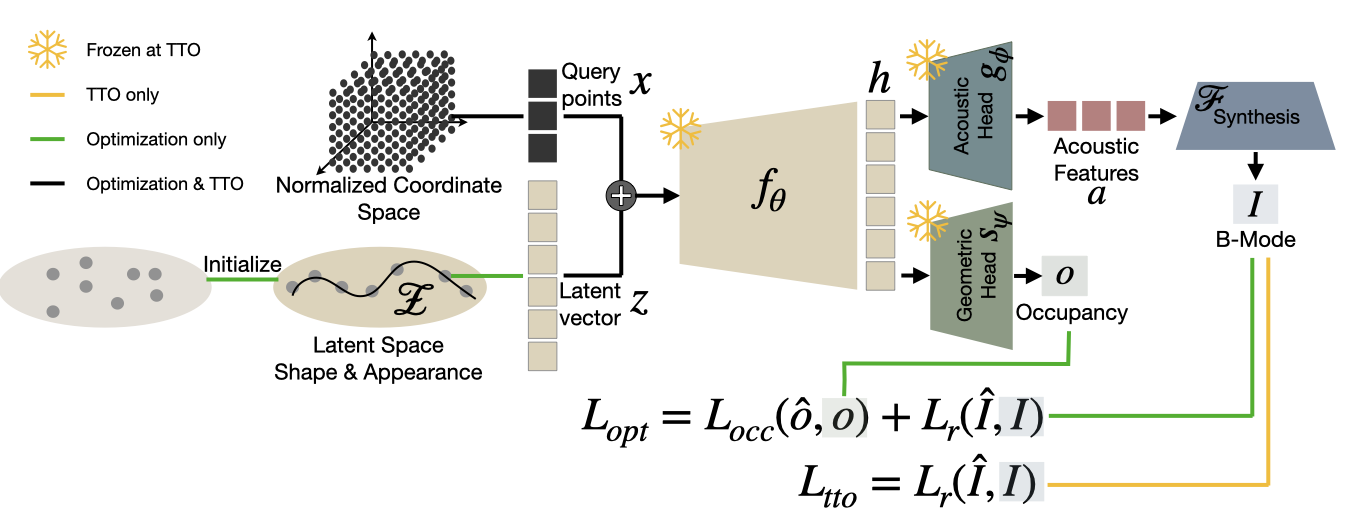} 
 \caption{\textbf{Proposed framework pipeline.} \textbf{Joint Training:} The network couples an acoustic head ($g_{\phi}$) and a geometric head ($s_{\psi}$) via a shared latent space ($\mathcal{Z}$) and backbone ($f_{\theta}$). Supervised by both B-mode synthesis ($\mathcal{L}_{photo}$) and ground-truth shape occupancy ($\mathcal{L}_{occ}$), the ray-based rendering inherently models acoustic shadowing.\textbf{Test-Time Optimization:} At inference, network weights are frozen. A novel latent code $\mathbf{z}^*$ is optimized directly from fragmented B-mode observations without shape labels ($\mathcal{L}_{tto}$). The optimized prior inherently extracts the complete 3D geometry $o(\mathbf{x} \mid \mathbf{z}^*)$.}
    \label{fig:main_figure}
\end{figure}
\subsection{Problem Formulation and Method Overview}
\label{subsec:methodoverview}
Given a sequence of partial B-mode ultrasound observations $\mathcal{D} = \{(I_k, T_k)\}_{k=1}^K$, with images $I_k$ and corresponding probe poses $T_k \in SE(3)$, our goal is to reconstruct the complete 3D vertebral geometry, defined as a continuous occupancy field $O: \mathbb{R}^3 \to [0, 1]$. We propose a NIR-based framework (Fig.~\ref{fig:main_figure}) that jointly models this geometry and acoustic properties. 

For a 3D coordinate $\mathbf{x}$ and a subject-specific latent code $\mathbf{z}$, a shared backbone network $f_{\theta}$ extracts a joint feature representation. This feature space is then processed by two distinct output heads. The acoustic head $g_{\phi}$ maps the joint features to a localized acoustic property vector $\mathbf{a}(\mathbf{x}) = g_{\phi}(f_{\theta}(\mathbf{x}, \mathbf{z}))$, where $\mathbf{a}(\mathbf{x}) = [\beta(\mathbf{x}), \sigma(\mathbf{x}), \mu(\mathbf{x})]^T$ represents acoustic reflection, scattering, and attenuation. 
Simultaneously, the occupancy head $s_{\psi}$ maps the same features to the continuous volume occupancy probability $o(\mathbf{x}) = s_{\psi}(f_{\theta}(\mathbf{x}, \mathbf{z}))$.

To bridge these 3D predictions with the 2D observations, a differentiable ray-based forward model synthesizes the expected B-mode intensity $\hat{I}_k(\mathbf{r}) = \mathcal{F}(\mathbf{r}, \mathbf{a})$. 
By explicitly calculating signal transmission $\mathcal{T}(\mathbf{x})$ along ray $\mathbf{r}$, the model gains an implicit awareness of occlusions, inherently capturing acoustic shadowing without explicit labels and learned latent prior completes the unobserved 3D shape $o(\mathbf{x} \mid \mathbf{z}^*)$.

\subsection{Joint Acoustic and Geometric Representation}
\label{subsec:latent_representation}
To capture the codependence between anatomical structure and its acoustic signature, we define a unified latent space $\mathcal{Z} \subset \mathbb{R}^d$. For any 3D coordinate $\mathbf{x} \in \mathbb{R}^3$ and subject latent code $\mathbf{z} \in \mathcal{Z}$, a shared backbone $f_{\theta}$ learns a joint structural prior:
$\mathbf{h}(\mathbf{x}) = f_{\theta}(\mathbf{x}, \mathbf{z})$.
This shared feature vector $\mathbf{h}(\mathbf{x})$ branches into two specialized heads. The acoustic head $g_{\phi}$ outputs physical tissue properties for wave propagation, $\mathbf{a}(\mathbf{x}) = g_{\phi}(\mathbf{h}(\mathbf{x})) = [\beta, \sigma, \mu]^T \ge 0$. Concurrently, the occupancy head $s_{\psi}$ yields the continuous surface occupancy, $o(\mathbf{x}) = s_{\psi}(\mathbf{h}(\mathbf{x})) \in [0, 1]$. By forcing both properties to decode from the shared representation $\mathbf{h}(\mathbf{x})$ conditioned on the same latent space $\mathcal{Z}$, the network intrinsically couples the spaces, projecting the acoustic optimization onto a manifold strictly constrained by the anatomical geometry.

\subsection{Physics-Aware Ray-Based Rendering}
\label{subsec:rendering}
To optimize the latent code $\mathbf{z}$ directly from B-mode images, we map the 3D acoustic field $\mathbf{a}(\mathbf{x})$ to the 2D image plane using the differentiable physical rendering formulation established in ultrasound NeRFs~\cite{wysocki2024ultra,wysocki2025ultron}.
Unlike standard optical neural rendering that integrates an entire ray into a single pixel, an ultrasound transducer ray $\mathbf{r}(t) = \mathbf{o} + t\mathbf{d}$ corresponds to a complete 1D scanline. Here, $\mathbf{o}$ is the origin at the transducer (derived from pose $T_k$), $\mathbf{d}$ is the unit directional vector, and $t \ge 0$ represents the axial depth along the ray.

The remaining acoustic energy, or transmission factor $\mathcal{T}(t)$, reaching depth $t$ is determined by integrating both the predicted attenuation field $\mu(\mathbf{x})$ and the accumulated reflection $\beta(\mathbf{x})$ from the transducer up to that specific depth. This explicitly models acoustic energy conservation: 
\begin{equation}
\mathcal{T}(t) = \exp \left( - \int_{0}^{t} \big( \mu(\mathbf{r}(s)) + \beta(\mathbf{r}(s)) \big) \, ds \right)
\end{equation}

The expected B-mode intensity $\hat{I}_k(\mathbf{r}(t))$ at depth $t$ is synthesized as the sum of the localized reflection $\beta(\mathbf{x})$ and scattering $\sigma(\mathbf{x})$, modulated by the remaining acoustic energy at that location:
\begin{equation}
\hat{I}_k(\mathbf{r}(t)) = \mathcal{T}(t) \big( \beta(\mathbf{r}(t)) + \sigma(\mathbf{r}(t)) \big)
\end{equation}

Crucially, when $\mathbf{r}(t)$ intersects highly reflective and attenuating structures such as vertebral bone at a specific depth ($t = t_{bone}$), the sum of $\mu(\mathbf{r}(t_{bone}))$ and $\beta(\mathbf{r}(t_{bone}))$ increases. This causes $\mathcal{T}(t)$ to exponentially decay toward zero for all subsequent depths $t > t_{bone}$. Because the synthesized intensity $\hat{I}_k$ at these deeper locations is multiplied by $\mathcal{T}(t) \approx 0$, the gradient flow to the acoustic parameters in the shadowed regions effectively vanishes during backpropagation. Consequently, the network inherently isolates unobserved anatomy without requiring explicit shadow segmentation, forcing the shape head to rely entirely on the learned latent prior to complete the occluded geometry.

\subsection{Optimization Strategy: Training and Inference}
Our framework operates in two distinct phases: global prior learning across the dataset and label-free test-time optimization (TTO) for unseen subjects. 

During the training phase, the network parameters $\{\theta, \phi, \psi\}$ and the dataset latent codes $\{\mathbf{z}_j\}_{j=1}^N$ are jointly optimized. The total training objective $\mathcal{L}_{opt}$ is a weighted sum of four components: a photometric loss $\mathcal{L}_{photo}$, a geometric occupancy loss $\mathcal{L}_{occ}$, an acoustic regularization $\mathcal{L}_{reg\_a}$, and a latent regularization $\mathcal{L}_{reg\_z}$:
\begin{equation}
\mathcal{L}_{opt} = \mathcal{L}_{photo} + \lambda_{occ} \mathcal{L}_{occ} + \lambda_{a} \mathcal{L}_{reg\_a} + \lambda_{z} \mathcal{L}_{reg\_z}
\end{equation}

The photometric loss $\mathcal{L}_{photo}$ measures the reconstruction quality of the synthesized B-mode images. To accurately capture both high-frequency structural details and low-frequency intensity variations, we formulate it as a weighted combination of the Structural Similarity Index Measure (SSIM) and L2 loss:
\begin{equation}
\mathcal{L}_{photo} = \alpha \big(1 - \text{SSIM}(\hat{I}, I)\big) + (1 - \alpha) \|\hat{I} - I\|_2^2
\end{equation}
where $I$ and $\hat{I}$ are the ground-truth and synthesized B-mode image patches, respectively, and $\alpha$ balances the two terms. 

The geometric shape is supervised via the Binary Cross-Entropy (BCE) loss between the predicted occupancy $o(\mathbf{x})$ and the ground-truth shapes $\hat{o}(\mathbf{x})$:
\begin{equation}
\mathcal{L}_{occ} = \sum_{\mathbf{x} \in \Omega} \text{BCE}\big(o(\mathbf{x}), \hat{o}(\mathbf{x})\big)
\end{equation}

To constrain the physical rendering, $\mathcal{L}_{reg\_a}$ regularizes the predicted acoustic parameter space. Finally, we apply standard L2 regularization to the latent codes, $\mathcal{L}_{reg\_z} = \|\mathbf{z}\|_2^2$, which enforces well-behaved latent manifold.

\noindent\textbf{During inference (TTO)} on an unseen subject, the trained network parameters $\{\theta, \phi, \psi\}$ are strictly frozen. We initialize a new latent vector $\mathbf{z}$ and optimize it without any geometric supervision. The TTO objective $\mathcal{L}_{tto}$ relies solely on the self-supervised photometric loss and the regularizers:
$$\mathcal{L}_{tto}(\mathbf{z}) = \mathcal{L}_{photo}(\hat{I}(\mathbf{z}), I) + \lambda_{a} \mathcal{L}_{reg\_a}(\mathbf{a}(\mathbf{x} \mid \mathbf{z})) + \lambda_{z} \|\mathbf{z}\|_2^2$$

The optimal latent code $\mathbf{z}^*$ is then obtained by minimizing this objective:
$$\mathbf{z}^* = \arg\min_{\mathbf{z}} \mathcal{L}_{tto}(\mathbf{z})$$

Because the frozen prior tightly couples the acoustic and geometric spaces, optimizing $\mathbf{z}^*$ strictly on the visible, unshadowed anatomy implicitly infers the global structure. The completed 3D vertebral geometry is then directly extracted by querying the occupancy field $o(\mathbf{x} \mid \mathbf{z}^*)$ across the spatial grid.

% \begin{figure}[htbp] % h=here, t=top, b=bottom, p=page
%     \centering
%     \includegraphics[width=\textwidth]{MICCAI2026-Latex-Template-4/gfx/compare_results.png} 
%     \caption{compare results}
%     \label{fig:compare_results}
% \end{figure}
% \section{Experiments and Results}
% \begin{figure}[htbp] % h=here, t=top, b=bottom, p=page
%     \centering
%     \includegraphics[width=\textwidth]{MICCAI2026-Latex-Template-4/gfx/compare_latent.png} 
%     \caption{compare latent}
%     \label{fig:compare_latent}
% \end{figure}
% \begin{figure}[htbp] % h=here, t=top, b=bottom, p=page
%     \centering
%     \includegraphics[width=\textwidth]{MICCAI2026-Latex-Template-4/gfx/interpolation_oscar.png} 
%     \caption{Interpolation Between Optimized Latent To Mean}
%     \label{fig:interpolation}
% \end{figure}
\section{Experiments and Results}
\subsection{Experimental Setup}
\subsubsection{Data}
Learning a rich shape prior requires extensive data. We therefore train and benchmark primarily on simulations, while utilizing physical phantoms to evaluate out-of-distribution generalization.
\textbf{Simulated Data:} Using ImFusion Suite~\footnote{ImFusion Suite, version 3.13.8} simulation~\cite{salehi2015patient}, we randomized acoustic properties to generate realistic B-mode images for 53 VerSe~\cite{sekuboyina2021verse} vertebrae. We simulated 5 sweeps per subject (perpendicular and $\sim 15^{\circ}$ tilted), producing 132 training, 10 validation, and 10 test sequences (500 images each with poses and occupancy labels). \textbf{Phantom Data:} We scanned three 3D-printed, unused in simulation, VerSe mesh embedded in a tissue-mimicking gelatine-paper pulp phantom~\cite{jiang2021autonomous}. 
A robot-mounted linear probe acquired the B-mode images, which were then registered to the original 3D model for ground truth comparison.
\subsubsection{Network Architecture \& Training \& TTO}
Our auto-decoder is an 8-layer MLP backbone (128 hidden units, ReLU) processing the concatenation of a 3D coordinate $\mathbf{x}$ and a 128-dimensional latent vector $\mathbf{z}$. We explicitly omit positional encoding to prevent overfitting to ultrasound speckle noise, yielding smoother surfaces. The backbone branches at the last layer into an acoustic head for physics-aware rendering and an occupancy head.
Initial latent codes are sampled from $\mathcal{N}(0, 10^{-3})$ and $L_2$-regularized. Models are trained for 10 epochs using the Adam optimizer with a learning rate of $10^{-4}$. We use validation set to select the best checkpoint and the optimal number of TTO iterations (4k for NISF, 6k for OSCAR). For TTO, the latent code is initialized by the mean latent code. 
\begin{figure}[t] % h=here, t=top, b=bottom, p=page
    \centering
    \includegraphics[width=0.9\textwidth]{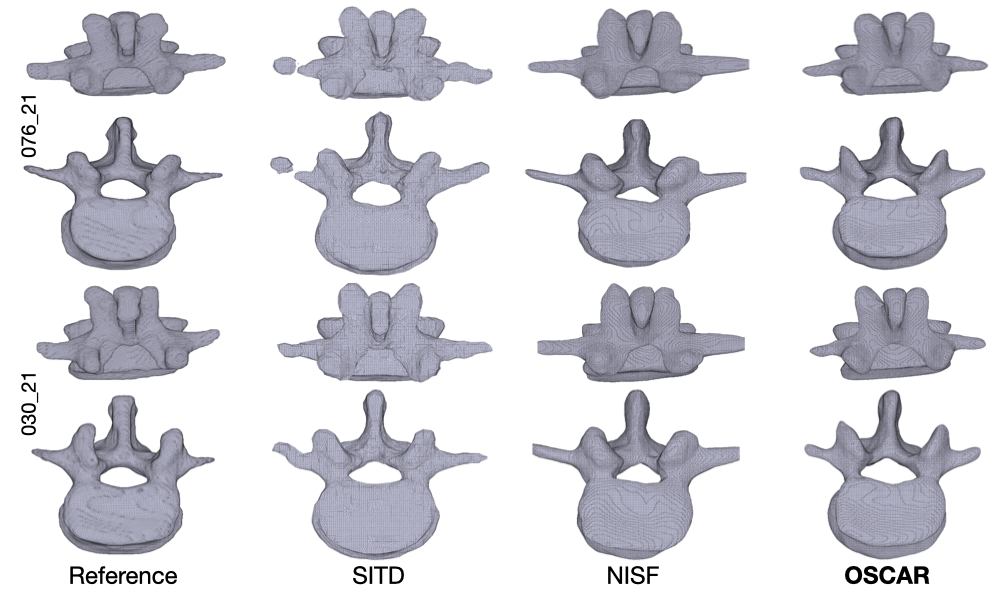} 
    \caption{\textbf{Visual analysis of reconstructed meshes.} OSCAR shows better structural awareness and accurate completion of invisible anatomy  compared to NISF (backbone) and SITD (which strictly requires segmentation inputs).}
    \label{fig:compare_results}
\end{figure}
\subsection{Evaluation}
% requires \usepackage{booktabs}
% for vertical text: \usepackage{graphicx}

\begin{table}[t]
    \centering
    \small
    \setlength{\tabcolsep}{6pt}
    \caption{Quantitative evaluation of reconstructed meshes compared to ground-truth for
    10 simulation and 3 phantom 
    test subjects.  The methods where trained on simulation data. We compare OSCAR against the baseline backbone (NISF) and SOTA approach (SITD). Performance is reported as Mean ± Std across three metrics: 95\% Hausdorff Distance (HD95), Bidirectional Chamfer Distance (CD), and F1-score
    % , and Earth Mover's Distance (EMD)
    .}
    \label{tab:model_results}
    \begin{tabular}{clcccc}
        \toprule
        & \textbf{Model} & \textbf{HD95(mm) $\downarrow$} & \textbf{CD $\downarrow$} & \textbf{F1 $\uparrow$} \\
        \midrule
        \multirow{3}{*}{\rotatebox{90}{\bf{Sim.}}} 
          & SITD &  $5.93 \pm 0.94$ & $119.87 \pm 17.04$ & $0.21 \pm 0.04$ \\
          & Nisf & $3.03 \pm 0.92$ & $88.38 \pm 14.43$ & $0.33 \pm 0.05$ \\
          & \textbf{Oscar} &\textbf{ 1.17 $\pm$ 0.38} & \textbf{56.57 $\pm$ 6.64} & \textbf{0.54 $\pm$ 0.06 } \\
        \midrule
        \multirow{3}{*}{\rotatebox{90}{\bf{{Phan.}}}} 
          & SITD &  $7.17 \pm 1.05$ & $104.66 \pm 20.50$ & $0.23 \pm 0.07$ \\
          & Nisf & $8.61 \pm 2.85$ & $146.98 \pm 22.57$ & $0.23 \pm 0.01$ \\
          & \textbf{Oscar} &\ $7.46 \pm 3.41$ & $105.34 \pm 20.99$ & $0.27 \pm 0.05$ \\
        \bottomrule
    \end{tabular}
\end{table}
% \begin{table}[b]
%     \centering
%     \small
%     \setlength{\tabcolsep}{6pt}
%     \caption{Quantitative evaluation of reconstructed meshes compared to ground-truth for 10 unseen test subjects after Test-Time Optimization (TTO). We compare OSCAR against the baseline NISF backbone and SOTA approach (SITD). Performance is reported as Mean ± Std across four metrics: 95\% Hausdorff Distance (HD95), Bidirectional Chamfer Distance (CD), F1-score, and Earth Mover's Distance (EMD).}
%     \label{tab:model_results}
%     \begin{tabular}{lccccc}
%         \toprule
%         \textbf{Model} & \textbf{HD95(mm) $\downarrow$} & \textbf{CD $\downarrow$} & \textbf{F1 $\uparrow$} & \textbf{EMD $\downarrow$} \\
%         \midrule
%         SITD &  $5.93 \pm 0.94$ & $119.87 \pm 17.04$ & $0.21 \pm 0.04$ & $867.25 \pm 101.82$ \\
%         Nisf & $3.03 \pm 0.92$ & $88.38 \pm 14.43$ & $0.33 \pm 0.05$ & $218.40 \pm 32.74$ \\
%         \textbf{Oscar} &\textbf{ 1.17 $\pm$ 0.38} & \textbf{56.57 $\pm$ 6.64} & \textbf{0.54 $\pm$ 0.06 }& \textbf{132.56 $\pm$ 17.13} \\
%         \bottomrule
%         %
%         \midrule
%         SITD &  $5.93 \pm 0.94$ & $119.87 \pm 17.04$ & $0.21 \pm 0.04$ & $867.25 \pm 101.82$ \\
%         Nisf & $3.03 \pm 0.92$ & $88.38 \pm 14.43$ & $0.33 \pm 0.05$ & $218.40 \pm 32.74$ \\
%         \textbf{Oscar} &\textbf{ 1.17 $\pm$ 0.38} & \textbf{56.57 $\pm$ 6.64} & \textbf{0.54 $\pm$ 0.06 }& \textbf{132.56 $\pm$ 17.13} \\
%         \bottomrule
%     \end{tabular}
% \end{table}
\subsubsection{Comparison with State of the Art}
We benchmark OSCAR against the current SOTA in B-mode ultrasound shape completion, Shape Completion in the Dark~\cite{gafencu2024shape} (SITD), and our backbone architecture, NISF~\cite{stolt2023nisf}. Notably, while NISF was originally designed strictly for segmentation, we demonstrate for the first time its inherent capacity for shape completion. 
As shown in Table \ref{tab:model_results}, our framework outperforms both baselines, yielding overall improvements of 80\% and 61\% in HD95, respectively. 
Furthermore, Figure \ref{fig:compare_results} provides a detailed quantitative and visual analysis of two representative samples from the test set. 
The gain over the baseline NISF proves that aligning acoustic and geometric features and optimizing strictly on visible anatomy better leverages the latent prior to complete occluded regions. 
Similarly, the improvement over SITD highlights that utilizing a continuous occupancy representation provides the structural awareness necessary for highly detailed shape recovery. In the phantom data, our label free method achieves comparable results to SITD which requires explicit labels.

% showing that physics-aware acoustic head helps to close sim-to-real domain gap.
\begin{figure}[b] % h=here, t=top, b=bottom, p=page
    \centering
    \includegraphics[width=0.9\textwidth]{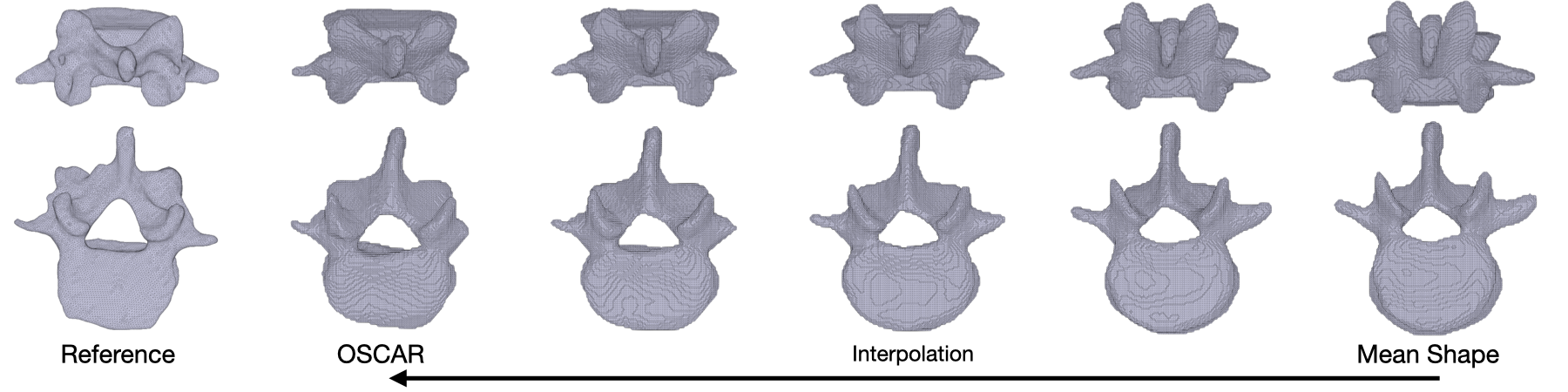} 
    \caption{\textbf{Latent interpolation} We morph the mean shape into a target shape by linearly interpolating their respective latent codes. The anatomically plausible intermediate states show that our framework learns a continuous and valid geometric prior.}
    \label{fig:latent}
\end{figure}
\subsubsection{Latent Prior}
To demonstrate that our learned prior captures a continuous anatomical manifold, we perform linear interpolation between the mean latent vector of the training set and a specific target shape. 
As shown in Figure \ref{fig:latent}, the interpolation is highly smooth, with the mean shape seamlessly morphing into the target geometry. Every intermediate step yields a valid vertebral structure, confirming that the latent space enforces strict anatomical plausibility.

\subsubsection{Acoustic Completion}
\begin{figure}[t] % h=here, t=top, b=bottom, p=page
    \centering
    \includegraphics[width=0.9\textwidth]{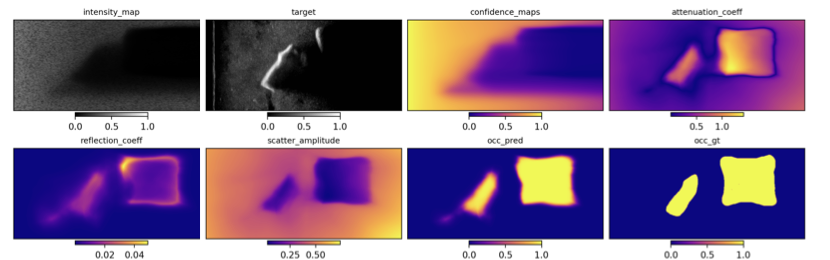} 
   \caption{\textbf{Bidirectional representation.} The acoustic space is accurately predicted by optimizing the latent code $\mathbf{z}$ using only the target shape, demonstrating a strong structural and acoustic coupling.}
    % \label{fig:acoustic_latent}
    \label{fig:bidirectional}
\end{figure}

To prove the shared latent space is bidirectional, we invert our optimization to update $\mathbf{z}$ using only the geometric shape. This successfully reconstructs the full acoustic space as visualized in Figure \ref{fig:bidirectional} on the example of the phantom image.
\begin{figure}[h] % h=here, t=top, b=bottom, p=page
    \centering
    \includegraphics[width=0.9\textwidth]{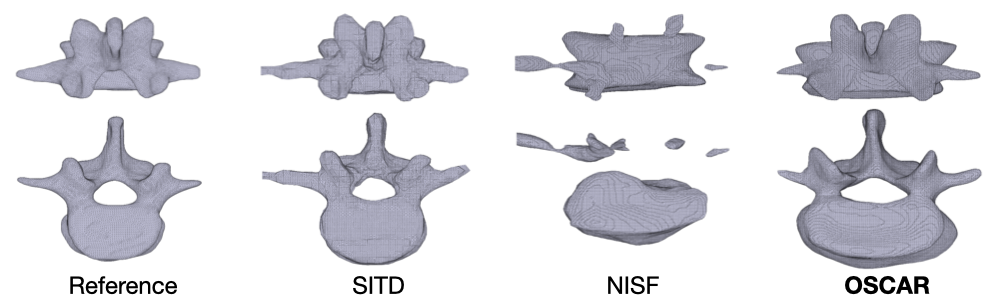} 
\caption{\textbf{Evaluation on phantom data.} Our framework accurately completes shapes from real B-modes despite the sim-to-real domain gap. In contrast, SITD avoids this OoD challenge only by requiring explicit point cloud labels.}
    \label{fig:phantom}
\end{figure}
\subsubsection{Out-of-distribution Shape Completion}
Because TTO optimizes solely on B-mode intensities, shifting distributions between simulated and real domains present a fundamental challenge. Nevertheless, Figure \ref{fig:phantom} demonstrates that our framework bridges this sim-to-real gap, accurately completing shapes on physical phantoms using only simulations during training..

\section{Conclusion}
In this work, we introduced OSCAR, a physics-aware framework for ultrasound-based 3D spinal shape completion. By integrating acoustic image formation and geometric occupancy into a coupled-latent implicit neural representation, OSCAR resolves multi-view inconsistencies and accurately reconstructs occluded anatomy. Crucially, while current SOTA methods rely on explicitly labeled surfaces or segmentation masks, our approach optimizes directly on raw B-mode image intensities. This direct optimization yields an 80\% improvement in the HD95 
% \textbf{CHECK NUMBERS} 
over the SOTA and enforces anatomical plausibility using an implicit prior. Preliminary evaluations on physical phantoms confirm that OSCAR can successfully translate these capabilities to real-world scans. Finally, the bidirectional capacity of our learned joint prior enables the synthesis of plausible acoustic spaces directly from known shapes, with the potential for more scalable registration and data synthesis in US imaging.
\newpage
\bibliographystyle{MICCAI2026-Latex-Template-4/splncs04}
\bibliography{MICCAI2026-Latex-Template-4/bib_shape}

\begin{thebibliography}{10}
\providecommand{\url}[1]{\texttt{#1}}
\providecommand{\urlprefix}{URL }
\providecommand{\doi}[1]{https://doi.org/#1}

\bibitem{burger2012real}
Burger, B., Bettinghausen, S., Radle, M., Hesser, J.: Real-time gpu-based ultrasound simulation using deformable mesh models. IEEE transactions on medical imaging  \textbf{32}(3),  609--618 (2012)

\bibitem{chen2024rocosdf}
Chen, H., Gao, Y., Zhang, S., Wu, J., Ma, Y., Zheng, R.: Rocosdf: row-column scanned neural signed distance fields for freehand 3d ultrasound imaging shape reconstruction. In: International Conference on Medical Image Computing and Computer-Assisted Intervention. pp. 721--731. Springer (2024)

\bibitem{duque2024ultrasound}
Duque, V.G., Marquardt, A., Velikova, Y., Lacourpaille, L., Nordez, A., Crouzier, M., Lee, H.J., Mateus, D., Navab, N.: Ultrasound segmentation analysis via distinct and completed anatomical borders. International Journal of Computer Assisted Radiology and Surgery  \textbf{19}(7),  1419--1427 (2024)

\bibitem{gafencu2025us}
Gafencu, M.A., Velikova, Y., Navab, N., Azampour, M.F.: Us-x complete: A multi-modal approach to anatomical 3d shape recovery. In: International Workshop on Shape in Medical Imaging. pp. 218--231. Springer (2025)

\bibitem{gafencu2024shape}
Gafencu, M.A., Velikova, Y., Saleh, M., Ungi, T., Navab, N., Wendler, T., Azampour, M.F.: Shape completion in the dark: completing vertebrae morphology from 3d ultrasound. International Journal of Computer Assisted Radiology and Surgery  \textbf{19}(7),  1339--1347 (2024)

\bibitem{jiang2021autonomous}
Jiang, Z., Li, Z., Grimm, M., Zhou, M., Esposito, M., Wein, W., Stechele, W., Wendler, T., Navab, N.: Autonomous robotic screening of tubular structures based only on real-time ultrasound imaging feedback. IEEE Transactions on Industrial Electronics  \textbf{69}(7),  7064--7075 (2021)

\bibitem{massalimova2025surgpointtransformer}
Massalimova, A., Liebmann, F., Jecklin, S., Carrillo, F., Farshad, M., F{\"u}rnstahl, P.: Surgpointtransformer: transformer-based vertebra shape completion using rgb-d imaging. Computer Assisted Surgery  \textbf{30}(1),  2511126 (2025)

\bibitem{mescheder2019occupancy}
Mescheder, L., Oechsle, M., Niemeyer, M., Nowozin, S., Geiger, A.: Occupancy networks: Learning 3d reconstruction in function space. In: Proceedings of the IEEE/CVF conference on computer vision and pattern recognition. pp. 4460--4470 (2019)

\bibitem{park2019deepsdf}
Park, J.J., Florence, P., Straub, J., Newcombe, R., Lovegrove, S.: Deepsdf: Learning continuous signed distance functions for shape representation. In: Proceedings of the IEEE/CVF conference on computer vision and pattern recognition. pp. 165--174 (2019)

\bibitem{salehi2015patient}
Salehi, M., Ahmadi, S.A., Prevost, R., Navab, N., Wein, W.: Patient-specific 3d ultrasound simulation based on convolutional ray-tracing and appearance optimization. In: International Conference on Medical Image Computing and Computer-Assisted Intervention. pp. 510--518. Springer (2015)

\bibitem{sekuboyina2021verse}
Sekuboyina, A., Husseini, M.E., Bayat, A., L{\"o}ffler, M., Liebl, H., Li, H., Tetteh, G., Kuka{\v{c}}ka, J., Payer, C., {\v{S}}tern, D., et~al.: Verse: a vertebrae labelling and segmentation benchmark for multi-detector ct images. Medical image analysis  \textbf{73},  102166 (2021)

\bibitem{stolt2023nisf}
Stolt-Ans{\'o}, N., McGinnis, J., Pan, J., Hammernik, K., Rueckert, D.: Nisf: Neural implicit segmentation functions. In: International Conference on Medical Image Computing and Computer-Assisted Intervention. pp. 734--744. Springer (2023)

\bibitem{vyas2025fit}
Vyas, K., Veeraraghavan, A., Balakrishnan, G.: Fit pixels, get labels: Meta-learned implicit networks for image segmentation. In: International Conference on Medical Image Computing and Computer-Assisted Intervention. pp. 194--203. Springer (2025)

\bibitem{de2025steerable}
de~Wilde, B., Rietberg, M.T., Lajoinie, G., Wolterink, J.M.: Steerable anatomical shape synthesis with implicit neural representations. In: International Conference on Medical Image Computing and Computer-Assisted Intervention. pp. 630--639. Springer (2025)

\bibitem{wysocki2024ultra}
Wysocki, M., Azampour, M.F., Eilers, C., Busam, B., Salehi, M., Navab, N.: Ultra-nerf: Neural radiance fields for ultrasound imaging. In: Medical Imaging with Deep Learning. pp. 382--401. PMLR (2024)

\bibitem{wysocki2025ultron}
Wysocki, M., Duelmer, F., Bal, A., Navab, N., Azampour, M.F.: Ultron: Ultrasound occupancy networks. In: International Conference on Medical Image Computing and Computer-Assisted Intervention. pp. 606--615. Springer (2025)

\bibitem{yang2024generating}
Yang, J., Sedykh, E., Adhinarta, J.K., Le, H., Fua, P.: Generating anatomically accurate heart structures via neural implicit fields. In: International Conference on Medical Image Computing and Computer-Assisted Intervention. pp. 264--274. Springer (2024)

\bibitem{yang2022implicitatlas}
Yang, J., Wickramasinghe, U., Ni, B., Fua, P.: Implicitatlas: learning deformable shape templates in medical imaging. In: Proceedings of the IEEE/CVF Conference on Computer Vision and Pattern Recognition. pp. 15861--15871 (2022)

\end{thebibliography}

\end{document}